\documentclass[runningheads]{llncs}
\usepackage{xcolor}
\usepackage{booktabs}
\usepackage{amsmath, amssymb}
\usepackage{float}
\usepackage{lmodern}
\usepackage{graphicx}
\usepackage{latexsym}
\usepackage{arydshln}
\usepackage{booktabs}
\usepackage{enumitem}
\usepackage{color}
\usepackage{graphicx}
\usepackage{booktabs}
\usepackage{amsmath}
\usepackage{amsfonts}
\usepackage{amssymb}
\usepackage{xcolor}
\usepackage{siunitx}
\usepackage{url}
\usepackage[misc]{ifsym}
\usepackage[ruled,linesnumbered]{algorithm2e}
\usepackage[T1]{fontenc}
\usepackage{lmodern}
\usepackage{textcomp}
\begin{document}
\title{Lag Selection for Univariate Time Series Forecasting using Deep Learning: An Empirical Study\thanks{This work was partially funded by projects AISym4Med (101095387) supported by Horizon Europe Cluster 1: Health, ConnectedHealth (n.º 46858), supported by Competitiveness and Internationalisation Operational Programme (POCI) and Lisbon Regional Operational Programme (LISBOA 2020), under the PORTUGAL 2020 Partnership Agreement, through the European Regional Development Fund (ERDF) and NextGenAI - Center for Responsible AI (2022-C05i0102-02), supported by IAPMEI, and also by FCT plurianual funding for 2020-2023 of LIACC (UIDB/00027/2020 UIDP/00027/2020)}}

\titlerunning{Lag Selection for Forecasting using Deep Learning}
%
\author{José Leites\inst{1}, Vitor Cerqueira\inst{2,3} \and
Carlos Soares\inst{2,3,4}}
\authorrunning{J. Leites et al.}

\institute{Faculdade de Ciências da Universidade do Porto, Porto, Portugal \\
\email{josemfleites@gmail.com}\and 
Faculdade de Engenharia da Universidade do Porto, Porto, Portugal \\
\email{\{csoares,vcerqueira\}@fe.up.pt}\and 
Laboratory for Artificial Intelligence and Computer Science (LIACC), Portugal \and
Fraunhofer Portugal AICOS, Portugal}

\maketitle              
\begin{abstract}

Most forecasting methods use recent past observations (lags) to model the future values of univariate time series.
Selecting an adequate number of lags is important for training accurate forecasting models.
Several approaches and heuristics have been devised to solve this task. However, there is no consensus about what the best approach is. Besides, lag selection procedures have been developed based on local models and classical forecasting techniques such as ARIMA. 
We bridge this gap in the literature by carrying out an extensive empirical analysis of different lag selection methods. We focus on deep learning methods trained in a global approach, i.e., on datasets comprising multiple univariate time series. 
The experiments were carried out using three benchmark databases that contain a total of 2411 univariate time series.
The results indicate that the lag size is a relevant parameter for accurate forecasts. In particular, excessively small or excessively large lag sizes have a considerable negative impact on forecasting performance.
Cross-validation approaches show the best performance for lag selection, but this performance is comparable with simple heuristics.

\end{abstract}

\section{Introduction}

Time series forecasting is an extensively studied problem with vast real-world applicability \cite{hyndman2018forecasting}.
Organizations from various domains leverage quantitative forecasts to drive their business planning.
The widespread interest in forecasting tasks led to the development of several forecasting methods. In recent benchmark competitions and datasets, deep learning approaches have achieved a state-of-the-art  forecasting performance when compared with classical approaches such as ARIMA \cite{oreshkin2019n,makridakis2018m4}.

Deep learning approaches model time series based on time delay embedding \cite{bontempi2013machine}. At each time step, the upcoming future values represent the output (target) variables. These are modeled based on the past recent observations or lags following an auto-regressive approach. Selecting an appropriate number of lags\footnote{This parameter is also referred to input size, lag size, or lookback window in the literature.} is an important step for training accurate forecasting models. An excessively large value may contain irrelevant information that that may result in overfitting. On the other hand, an excessively small lag size may miss patterns and long-range dependencies.

There are several methods in the literature to estimate the number of lags. These include approaches based on the partial auto-correlation function (PACF) \cite{hyndman2018forecasting}, false nearest neighbors \cite{kennel1992determining}, or heuristics based on the task parameters \cite{bandara2020lstm}.
However, there is no consensus about what the best approach is. Besides, most of these have been developed for local methods based on classical forecasting techniques such as ARIMA. A local forecasting model is fit using only the historical data of a given input time series \cite{januschowski2020criteria}.
It is not clear how these approaches should be applied for datasets involving multiple univariate time series that are typically used to train global forecasting models based on deep learning.

We bridge this gap in the literature by contributing with an extensive empirical analysis of the number of lags and its impact on forecasting performance. We focus on deep learning and a global forecasting setting, where a neural network is trained using a collection of univariate time series. Specifically, we use NHITS \cite{challu2023nhits}, a state-of-the-art deep learning method for univariate time series forecasting.

The experiments encompass three benchmark time series databases that contain a total of 2411 time series and 321734 observations.
The results indicate that the number of lags has a noticeable impact on forecasting performance. We found that too small or too large values can reduce performance considerably.
In terms of lag selection procedures, an approach based on cross-validation led to the best overall performance. However, we also found that an heuristic and a PACF-based approach show a comparable performance.
The code for reproducing the experiments is available online\footnote{\url{https://github.com/jmleites/Lag_selection/}}.

\section{Background}\label{sec:background}

This section introduces the main topics related to this work. First, we formalize the forecasting predictive task using univariate time series (Section \ref{sec:2.1}). We emphasize global forecasting models which is the standard approach for training deep neural networks.
In Section \ref{sec:2.2}, we describe a few neural network architectures for modeling time series. Finally, we describe several approaches for selecting the number of lags (Section \ref{sec:2.3}).

\subsection{Univariate time series forecasting}\label{sec:2.1}

A univariate time series is a set of time-ordered numeric values $Y = \{y_1, y_2, \dots,$ $y_t \}$, where $y_i \in \mathbb{R}$ is the value of the $i$-th observation and $t$ is the length of the time series. 
We address univariate time series forecasting tasks. The goal is to predict the future value of observations of the time series, $y_{t+1}, \ldots, y_{t+H}$, where $H$ is the forecasting horizon.

One standard approach for time series forecasting involves transforming the time series from a sequence of values based on Takens time delay embedding theorem \cite{Takens1981}. Time delay embedding works by reconstructing a time series into the Euclidean space by applying sliding windows. This results in a dataset $\mathcal{D}=\{$X$, $y$\}^t_{p+1}$ where $y_i$ represents the $i$-th observation and $X_i \in \mathbb{R}^p$ is the $i$-th corresponding set of $p$ lags: $X_i = \{y_{i-1}, y_{i-2}, \dots, y_{i-p} \}$. 
According to Takens theorem, the reconstructed time series preserves the properties of the original one, provided an adequate transformation parameters \cite{manabe2007novel}. In effect, the forecasting task is framed as a multiple regression problem where $p$ lags are used as explanatory variables and future values are the target variables.

\subsection{Forecasting with deep learning}\label{sec:2.2}

Several forecasting tasks involve multiple univariate time series. For instance, the problem of forecasting the sales of multiple retail products. Let $\mathcal{Y} = \{Y_1, Y_2, \dots, Y_n\}$ denote a set of unordered time series, where $Y_j$ is the $j$-th time series and $n$ is the number of time series in the collection.
In general, forecasting approaches fall into a local or global methodology. Classical forecasting techniques, such as ARIMA, are local methods \cite{januschowski2020criteria} as they build an forecasting model for each time series in $\mathcal{Y}$.
On the other hand, global methods train a single forecasting model using multiple time series as input. Global forecasting models have shown state-of-the-art forecasting performance in benchmark competitions \cite{makridakis2018m4,godahewa2021ensembles}. The key motivation for using a global approach is that a model can learn useful patterns across different time series.

The time delay embedding approach described in Section \ref{sec:2.1} can be applied for datasets involving multiple time series. Particularly, the training data $\mathcal{D}$ is a concatenation of the individual datasets concerning the individual time series: $\mathcal{D} = \{\mathcal{D}_1, \dots,  \mathcal{D}_n\}$, where $\mathcal{D}_j$ is the $j$-th reconstructed time series based on time delay embedding.

Machine learning algorithms, such as artificial neural networks, typically follow a global methodology.
Several neural networks have recently achieved state-of-the-art forecasting performance. ES-RNN is a popular architecture based on LSTM that won the M4 forecasting competition \cite{smyl2020hybrid}. While architectures based on recurrent neural networks, such as the LSTM, are popular for time series, transformers \cite{zhou2021informer} or methods based on multi-layer perceptrons (MLPs) are also competitive \cite{challu2023nhits}.
NHITS (Neural Hierarchical Interpolation for Time Series Forecasting) \cite{challu2023nhits} is a recently proposed architecture based on blocks of MLPs with residual connections. This approach has been shown to provide better forecasting performance than several other neural networks as well as classical forecasting techniques \cite{challu2023nhits,cerqueira2024fly}. Besides a competitive forecasting performance, NHITS is also significantly more efficient computationally \cite{challu2023nhits}.

\subsection{Lag selection}\label{sec:2.3}

The approach for modeling time series put forth in the previous section involves the selection of the input size $p$. This parameter denotes the number of lags to use, i.e., how far back in time the model should go to predict the value of upcoming observations.

Lag selection is often carried out according to PACF values \cite{hyndman2018forecasting}.
PACF quantifies the correlation of a time series with its lags. Unlike the standard auto-correlation function (ACF), in the PACF the correlation takes into account the correlation at shorter lag. The examination of the PACF plot can be used to guide the lag selection process. For example, select a number of lags at which the PACF falls below a significant value. This choice can also be driven using information criterion methods such as the Akaike Information Criterion (AIC) or the Bayesian Information Criterion (BIC) \cite{hyndman2018forecasting}.
The configuration of ARIMA, a well-established classical forecasting technique, is often driven by this approach \cite{box2015time}. 

The False Nearest Neighbours (FNN) \cite{kennel1992determining} is another commonly used approach for selecting the number of lags. FNN performs a nearest neighbours analysis for increasing lags. The intuition for this method is that, with too low number of lags, many of the nearest neighbours will be false or spurious \cite{kennel1992determining}. These false neighbors disappear as we increase the number of lags towards an optimal value. FNN is typically applied for non-linear time series analysis, e.g.~\cite{lukoseviciute2010evolutionary}. This method includes a tolerance threshold for the allowed ratio of false neighbors. A smaller tolerance level will lead to a larger estimated lag size.

Evolutionary algorithms have also been studied for lag selection, e.g. \cite{parras2014short,bouktif2018optimal}.
Another approach is to use principal component analysis (PCA) to circumvent the lag selection task \cite{albano1988singular}. The idea is to reconstruct the time series with an arbitrarily large lag size and apply PCA to this data. Then, a model is trained using the principal components as explanatory variables. One can also select the number of lags using simple heuristics based on the forecasting problem. For example, Bandara et al. \cite{bandara2020lstm} suggest selecting a number of lags based on the forecasting horizon and the sampling frequency of the time series. The idea is to take the maximum value between the forecasting horizon and the frequency (e.g. 12 for monthly time series), and then multiply the result by a factor of 1.25.

The lag selection process can be framed as a model selection problem that can be solved using cross-validation \cite{cerqueira2023model}. The idea is to evaluate a method using different lags and select the configuration that maximizes forecasting performance. The performance estimates can be coupled with information criteria such as AIC to penalize models with an increasing number of lags \cite{shibata1976selection}.

\section{Materials and Methods}\label{sec:materials}

This section details the materials and methods used in the experiments. These include the datasets (Section \ref{sec:data}), evaluation methodology (Section \ref{sec:evalmethod}), and forecasting and lag selection methods (Section \ref{sec:methods}).

\subsection{Data}\label{sec:data}

The experiments are carried out using three time series databases that were part of different forecasting competitions: M1 \cite{makridakis1982accuracy}, M3 \cite{makridakis2000m3}, and Tourism \cite{athanasopoulos2011tourism}. These datasets are well-established in the literature for the evaluation of univariate forecasting methods.
We focus the study on monthly time series in the interest of conciseness.
Table \ref{tab:data} presents a brief summary of these datasets.
 
\begin{table}
\caption{Average number of values, number of time series, and number of observations of each database.}
\label{tab:data}
\begin{tabular}{lrrr}
\toprule
Database & Avg. length & \# time series & \# observations \\
\midrule
M1 Monthly & 72.7 & 617 & 44892 \\
M3 Monthly & 117.3 & 1428 & 167562 \\
Tourism Monthly & 298.5 & 366 & 109280 \\
\midrule
Total & - & 2411 & 321734 \\
\bottomrule
\end{tabular}
\end{table}

\noindent The forecasting horizon $H$ is set to 18 for the M3 and Tourism datasets. This parameter is set to 8 on M1 dataset due to constraints related to the size of the time series composing this dataset. We follow a multi-input multi-output approach for multi-step ahead forecasting \cite{cerqueira2024instance} as neural networks can natively handle multiple target variables in a single model.

\subsection{Evaluation methodology}\label{sec:evalmethod}

Regarding the data partitioning process, we apply an holdout approach by individual time series. Specifically, the test set is composed of the last $H$ observations of each time series, and the remaining data is used for training the model. 
Lag selection based on performance estimates (detailed in Section \ref{sec:methods}) is done using a validation set. This process involves further splitting the training set into two parts, where the validation set is created in a similar fashion as the test set. It is composed of the $H$ observations of each time series before the $H$ testing observations, following a time series cross-validation procedure \cite{cerqueira2023model}. 

We use SMAPE (symmetric mean absolute percentage error) to evaluate performance, which is defined as follows.
\begin{equation}
    \text{SMAPE} = \frac{100\%}{n} \sum_{i=1}^{n} \frac{|\hat{y}_i - y_i|}{(|\hat{y}_i| + |y_i|)/2}
\end{equation}

\noindent where $\hat{y}_i$, and $y_i$ are the forecast and actual value for the $i$-th instance, respectively. SMAPE is typically used to quantify forecasting performance, for example in the M4 competition \cite{makridakis2018m4}.

\subsection{Methods}\label{sec:methods}

\subsubsection{Forecasting methods.}

This work focuses on global forecasting methods, particularly on deep learning approaches. In the interest of conciseness, we test a single architecture, specifically NHITS, which we described in Section \ref{sec:2.2}. As mentioned, this architecture has achieved state-of-the-art forecasting performance with better computational scalability relative to other neural networks.

The training procedure is the following. We train a model based on NHITS for each dataset in Table \ref{tab:data}, using the default implementation available on PyTorch-based neuralforecast\footnote{\url{https://nixtlaverse.nixtla.io/neuralforecast/models.nhits.html}}. The hyperparameters include three stacks, each with one MLP. Each MLP comprises two hidden layers, each with 512 units (with rectified linear activation).
We train the model for 1500 training steps based on an ADAM optimization and a learning rate equal to 0.001. The training process also includes model checkpointing and early stopping (with 50 patience steps).
Besides NHITS, we also include a seasonal naive forecasting technique as baseline. This method sets the prediction for future values of time series equal to the last known observation of the same period.

\subsubsection{Lag selection methods.}\label{sec:lagselection}

We include the following lag selection methods in the experiments:

\begin{itemize}
    \item \texttt{FNN\@0.001}, \texttt{FNN\@0.01}: The FNN method using a tolerance for false neighbors of 0.001 and 0.01, respectively. This method estimates the number of lags for individual univariate time series in a given database $\mathcal{Y}$. In effect, we need to aggregate the individual estimations into a final estimated number of lags to train a forecasting model. We carry out this process using the average of individual estimations. More precisely, We average all estimated number of lags and then, take the ceiling of this average to obtain an integer global estimation for the number of lags;

    \item \texttt{PACF\@0.001}, \texttt{PACF\@0.01}: The PACF method using a significance value threshold of 0.001 and 0.01, respectively. For each time series, we select the number of lag at which the PACF value falls below the respective threshold. We aggregate the results across all time series using the same process as described above for \texttt{FNN\@0.001} and \texttt{FNN\@0.01}.

    \item \texttt{Frequency}: An heuristic that sets the number of lags to be equal to the frequency of the time series to cover a complete seasonal cycle. Since we only cover monthly time series in this work, the estimated number of lags is constant 12;

    \item \texttt{2*Frequency}: Similar to \texttt{Frequency}, but covering two seasonal cycles. This leads to a constant number of 24 estimated lags;

    \item \texttt{Horizon}: Another heuristic that set the number of lags to the forecasting horizon;

    \item \texttt{Bandara}: The heuristic by Bandara et al. \cite{bandara2020lstm}, which we describe in Section \ref{sec:2.3};

    \item \texttt{Previous}: A baseline that sets the number of lags to 1. In effect, future values are modeled based on the last known value, resembling a naive forecasting technique \cite{hyndman2008automatic};

    \item \texttt{CV}: A cross-validation approach for estimating the number of lags. This method works by evaluating different lag values using a validation set (described in the previous section). The estimated number of lags is the one that maximizes forecasting performance according to SMAPE;

    \item \texttt{AR}: A variant of \texttt{CV} that uses the average rank to estimate the number of lags \cite{brazdil2000comparison}. More precisely, we compute the SMAPE in the validation set of all time series in a given dataset. Then, we compute the average rank of each lag using these scores. The selected number of lags is the one with the best average rank. Since the rank is non-parametric, it is more robust to outliers.

\end{itemize}

\noindent In the lag selection process, we evaluate all possible number of lags up to a maximum of 120. Since the datasets contain monthly time series, this is equivalent of 10 years of data.

\begin{table}
\centering
\caption{Number of selected lags by each method and for each dataset.}
\label{tab:selectedlags}
\begin{tabular}{lrrr}
\toprule
Method & M3 & Tourism & M1 \\
\midrule
\texttt{AR} & 63 & 43 & 58 \\
\texttt{Bandara} & 23 & 23 & 15 \\
\texttt{CV} & 17 & 43 & 66 \\
\texttt{FNN@0.001} & 5 & 7 & 4 \\
\texttt{FNN@0.01} & 5 & 6 & 4 \\
\texttt{Frequency} & 12 & 12 & 12 \\
\texttt{Frequency*2} & 24 & 24 & 24 \\
\texttt{Horizon} & 18 & 18 & 8 \\
\texttt{PACF@0.001} & 44 & 77 & 29 \\
\texttt{PACF@0.01} & 24 & 29 & 19 \\
\texttt{Previous} & 1 & 1 & 1 \\
\bottomrule
\end{tabular}
\end{table}

\section{Experiments}

This section presents the experiments carried out to analyse the number of lags in univariate time series forecasting problems using global models based on deep learning.
We address the following research questions:
\begin{itemize}

    \item RQ1: How does the performance of NHITS, a state-of-the-art deep learning forecasting method, vary with the number of lags for univariate time series forecasting?

    \item RQ2: How does NHITS compare with seasonal naive, a baseline classical forecasting technique?

    \item RQ3: What is the best approach for selecting the number of lags for univariate time series forecasting using deep learning? Specifically, we focus on time series dataset involving several time series where a model is trained with a global approach.

\end{itemize}

Table \ref{tab:selectedlags} details the number of selected lags by each method on each dataset. The estimated value varies across each method and dataset. 

\subsection{Results}

\begin{figure}[h]
    \centering
    \includegraphics[width=\textwidth]{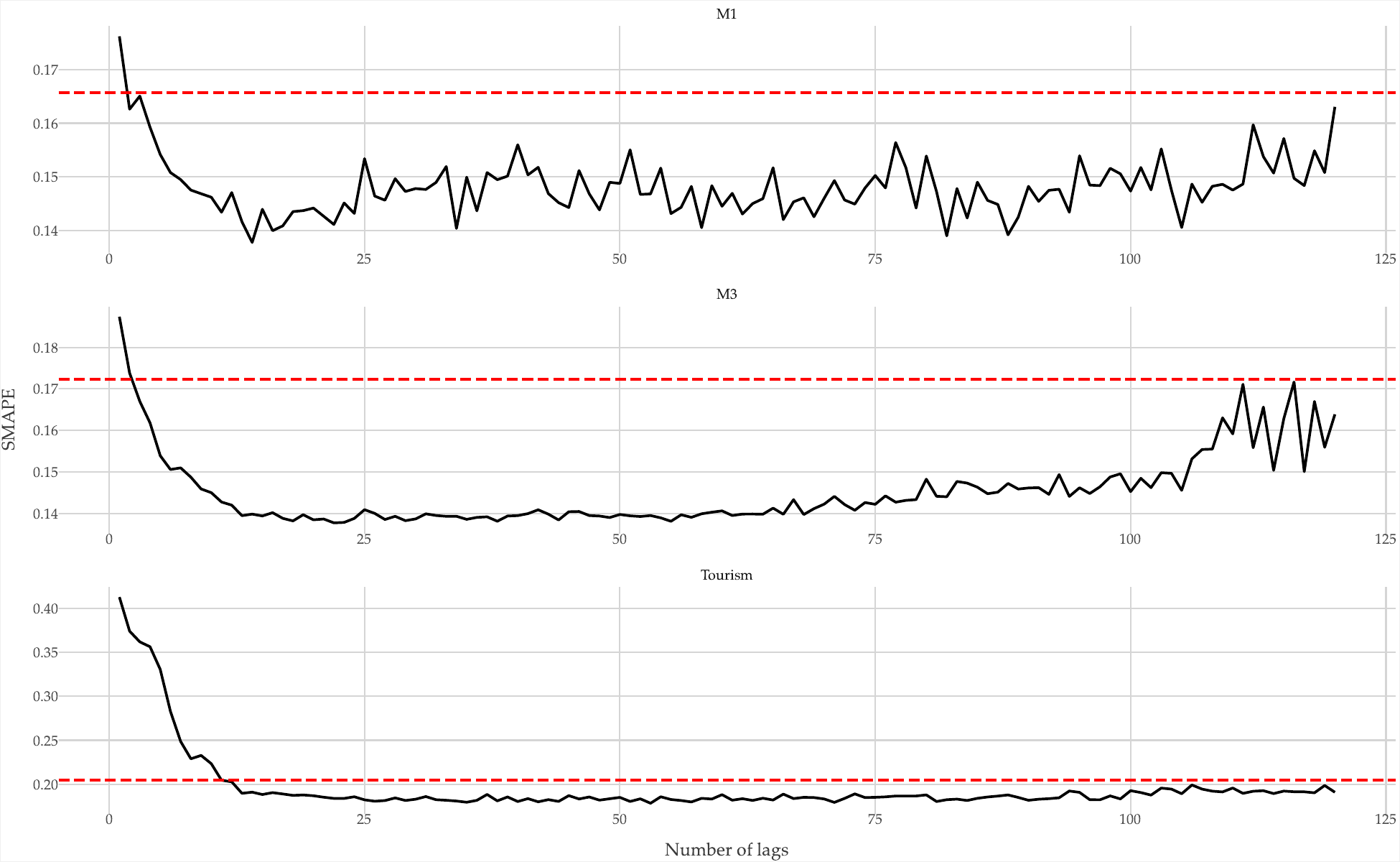}
    \caption{SMAPE scores for an increasing number of lags across each dataset. The dashed red line denotes the performance of the seasonal naive baseline.}
    \label{fig:bars_main}
\end{figure}

We start by analysing the performance of NHITS for an increasing number of lags. The results are shown in Figure \ref{fig:bars_main}, which illustrates the SMAPE scores for lag size and for each dataset. The dashed red line denotes the performance of the seasonal naive baseline.

Overall, the number of lags has a noticeable impact on forecasting performance. In particular, a very small lag size reduces forecasting performance to a level below the seasonal naive baseline. In two of the datasets (M3, M1), a very large lag size also leads to a decrease in forecasting performance. 
When the lag size is set in between these extremes, the impact of the number of lags is less significant.

\begin{figure}[!ht]
    \centering
    \includegraphics[width=\textwidth]{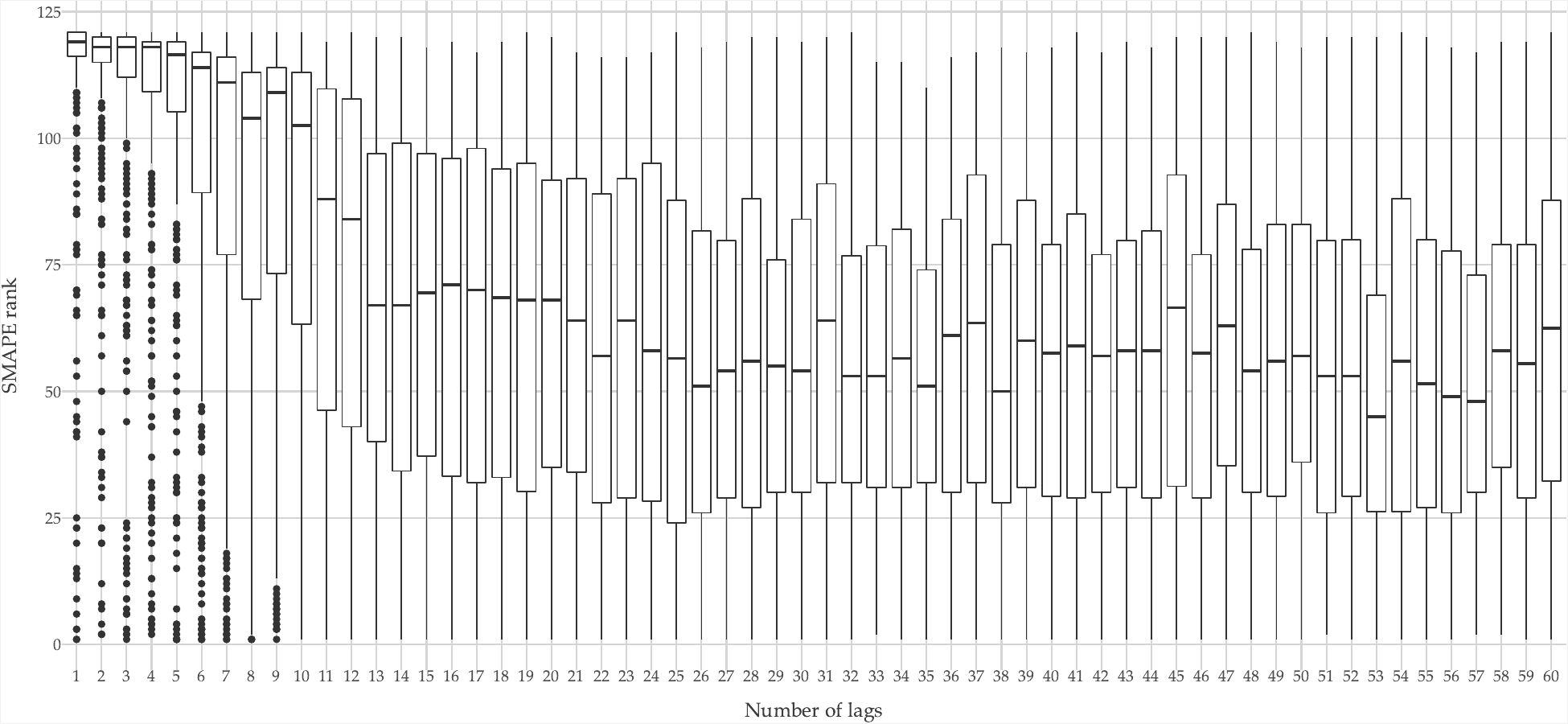}
    \caption{Rank distribution of each lag across all time series in the M3 dataset. For visualization purposes, we truncate the analysis to the first 60 lags.}
    \label{fig:bars_ranks}
\end{figure}

We explore the results further in Figure~\ref{fig:bars_ranks}, which shows the rank distribution of each lag across the time series in the M3 dataset. This dataset was selected arbitrarily, and we found similar results in the other two datasets. A given lag has a rank of 1 in a given time series if it shows the best performance on that problem. Overall, different lag sizes show better performance in different time series. This result shows that a single lag size is not the optimal solution to all time series in a dataset.

\begin{table}
\caption{SMAPE of each lag selection method across each dataset, and the respective average. Methods are ordered by average score. Bold font denotes the best performance on the respective dataset.}
\label{tab:results}
\begin{tabular}{lllll}
\toprule
 & M3 & Tourism & M1 & Average \\
\midrule
\texttt{AR} & 0.1399 & 0.1824 & \textbf{0.1406} & \textbf{0.1543} \\
\texttt{CV} & 0.1388 & 0.1824 & 0.1421 & 0.1544 \\
\texttt{PACF@0.01} & 0.1388 & \textbf{0.1814} & 0.1438 & 0.1547 \\
\texttt{Bandara} & \textbf{0.1379} & 0.1837 & 0.144 & 0.1552 \\
\texttt{Frequency*2} & 0.1388 & 0.1856 & 0.1432 & 0.1559 \\
\texttt{PACF@0.001} & 0.1384 & 0.1865 & 0.1473 & 0.1574 \\
\texttt{Horizon} & 0.1382 & 0.1872 & 0.1476 & 0.1577 \\
\texttt{Frequency} & 0.142 & 0.2026 & 0.1471 & 0.1639 \\
\texttt{FNN@0.001} & 0.1539 & 0.2485 & 0.1592 & 0.1872 \\
\texttt{FNN@0.01} & 0.1539 & 0.2827 & 0.1592 & 0.1986 \\
\texttt{Previous} & 0.1874 & 0.4129 & 0.1761 & 0.2588 \\
\bottomrule
\end{tabular}
\end{table}

Table \ref{tab:results} shows the performance of different lag selection methods in different datasets, along with the combined average. 
Overall, the top performing approaches share a comparable performance. \texttt{AR} shows the best performance, followed by \texttt{CV} and \texttt{PACF@0.01}.
The baseline \texttt{Previous} that uses a single lag shows the worst forecasting performance. The approaches based on false neighbors (\texttt{FNN@0.001} and \texttt{FNN@0.01}) estimate a small number of lags and also under-perform relative to other methods.
In the PACF-based approach, setting a significance value to 0.01 (\texttt{PACF@0.01}) shows better performance than a more strict value of 0.001 (\texttt{PACF@0.001}).

\section{Discussion}\label{sec:discussion}


The results of the experiments provide evidence regarding the research questions put forth. Overall, the lag size as an impact on the forecasting performance of NHITS (RQ1). For all datasets, NHITS shows a poor performance if the lag size is set too small. As the lag size increases, forecasting performance improves and becomes stable. In two of the datasets, the performance decreases again for large lag sizes.
Overall, avoiding a very small or a very large number of lags is an important aspect when setting this parameter.
With a sufficiently large lag size, NHITS outperforms the seasonal naive baseline (RQ2).

Finally, we also compared different approaches for automatic lag selection. Some of these approaches were originally designed for forecasting models trained on a single time series (local models). We adapted the application of these methods to datasets involving multiple time series (c.f. Section \ref{sec:lagselection}). 
\texttt{AR} and \texttt{CV}, two approaches based on cross-validation, presented the best performance. However, \texttt{PACF@0.01} and two heuristics (\texttt{Bandara} and \texttt{Frequency*2}) showed a comparable performance with these.
This result is important because cross-validation is considerably more costly computationally than computing PACF values (for \texttt{PACF@0.01}) or heuristics (RQ3).

We explored the results in terms of variability of the best lag size. We found that different lag sizes are better for different time series in a dataset. In effect, training a global forecasting model using a single lag size leads to sub-optimal results overall. A potentially interesting research direction is to build forecasting models that uses different lag sizes. 
For example, previous works \cite{cerqueira2017dynamic,oreshkin2019n} have built ensembles of forecasting models, where each model uses different lag sizes to foster diversity in the ensemble.

Lag selection can be framed as a particular instance of feature selection where the explanatory variables are ordered in time. In this context, another potentially promising research direction is to adapt state-of-the-art feature selection approaches for lag selection, for example, RReliefF \cite{robnik2003theoretical}.

\section{Conclusions}

This work presents an empirical analysis of the number of lags for univariate time series forecasting tasks. We focus on deep learning approaches, specifically the state-of-the-art method NHITS \cite{challu2023nhits}. We also focus on a global setting where the neural network is trained with multiple time series. In terms of data, we used three benchmark databases with monthly univariate time series.

Overall, we found that the lag size has an impact on forecasting performance. 
Avoiding a too small lag size is critical for an adequate forecasting performance, and an excessively large lag size can also reduce performance.
Cross-validation approaches for lag selection lead to the best performance. However, lag selection based on PACF or based on heuristics show a comparable performance with these.

Another interesting result is that different lag sizes show better performance in distinct time series in the dataset. In future work, we will explore this result to improve the performance of global forecasting models based on deep learning.

\bibliographystyle{splncs04}

\end{document}